\documentclass[11pt]{article}

\usepackage[preprint]{acl}

\usepackage{times}
\usepackage{latexsym}

\usepackage[T1]{fontenc}

\usepackage[utf8]{inputenc}

\usepackage{microtype}

\usepackage{inconsolata}

\usepackage{graphicx}

\title{DPH Parser: A Bottom-Up Grammar-Driven Parser for Joint Constituency and Dependency Analysis }

\author{Hussein Ghaly \\
  Independent Researcher \\
  New York, New York \\
  \texttt{hmghaly@gmail.com} }

\begin{document}
\maketitle
\begin{abstract}
This paper presents Dependency-Phrase Hierarchy Parser (DPH Parser), a grammar-driven bottom-up unsupervized parsing framework inspired by Generalized Phrase Structure Grammar (GPSG) and Head-driven Phrase Structure Grammar (HPSG). The parser incrementally constructs constituency structures using a compact inventory of feature-based syntactic rules while deriving dependency relations through explicit head annotations. The system combines probabilistic POS tagging, recursive phrase projection, and weighted parse hypotheses to process realistic and partially noisy text input. Unlike purely neural and data-driven parsers, the resulting syntactic derivations remain explicitly interpretable.

We evaluated parser performance on English corpora from the Universal Dependencies (UD) project using Unlabeled Attachment Score (UAS) as the main parsing metric, comparing the outcomes against Stanza and spaCy parsers. For a small inventory of syntactic rules, DPH parser achieved UAS values of 53.32\% \& 52.58\% (UD Devset/Testset respectively). For the same data, Stanza achieved 89.12\% \& 88.67\% while spaCy achieved 56.91\% and 58.59\%. Although the current system does not yet approach the accuracy of modern neural parsers, the results demonstrate the feasibility of applying transparent rule-based bottom-up parsing to realistic treebank data while jointly producing constituency and dependency structures.

\end{abstract}

\section{Introduction}

Natural language parsing remains a fundamental problem in computational linguistics despite the recent success of large neural language models (LLMs). Modern transformer-based systems often achieve strong downstream performance without explicit syntactic representations, and contemporary dependency parsers routinely achieve high attachment accuracy on benchmark corpora. Nevertheless, these systems typically encode syntactic structure implicitly within distributed representations, making their structural decisions difficult to interpret, constrain, or modify explicitly. This is particularly important for sentences with ambiguity, with multiple possible underlying structures. For example, in a prompt such as: “generate a picture of men and women wearing hats”, do both men and women wear hats, or only women? Therefore, even for more advanced LLM and neural-based NLP applications, it is still important to recognize ambiguity, while maintaining proper syntactic derivation of the underlying structure.

The goal of the current work is not to replace modern neural parsers, but rather to investigate whether a compact inventory of linguistically motivated syntactic rules can support a practical parsing framework that incrementally constructs interpretable phrase structures while simultaneously preserving dependency information.

The proposed grammar formalism is primarily inspired by Generalized Phrase Structure Grammar (GPSG) \cite{gazdar_generalized_1985}  and Head-driven Phrase Structure Grammar (HPSG) \cite{pollard_head-driven_1994}. In contrast to transformational approaches of generative syntax (starting with \citealt{chomsky_syntactic_1957}), these frameworks attempt to represent long-distance dependencies and syntactic constraints through feature propagation mechanisms rather than explicit movement operations. In particular, GPSG introduced slash features for representing constituents with missing arguments (e.g. the object of the verb “eat” in “What did you eat X?”), while HPSG emphasized typed feature structures and explicit head propagation. These ideas provide useful mechanisms for constructing phrase structures while simultaneously preserving dependency relations between words. 
\begin{figure}
    \centering
    \includegraphics[width=1\linewidth]{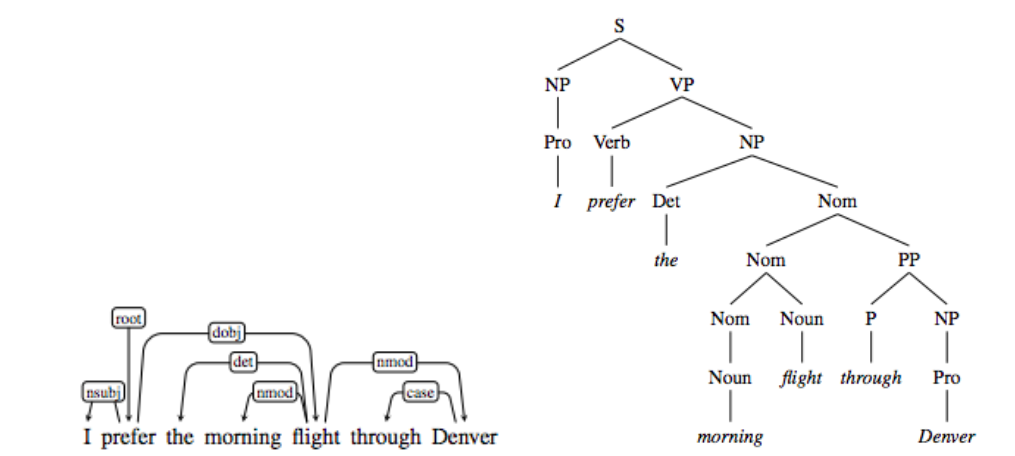}
    \caption{Dependency and Constituency Structures }
    \label{fig:Dependency and Constituency Structures }
\end{figure}

 This formalism addresses a common issue for modern parsing systems, which often distinguish between constituency parsing (producing hierarchical phrase structure), and dependency parsing (producing binary head-dependent relations between words). Both representations are interrelated, and each can be converted into the other, but each has a certain focus and provides certain information about the structure \cite{xia_converting_2001}. While dependency parsing dominates many contemporary NLP pipelines because of its efficiency and compatibility with downstream applications \cite{nivre_inductive_2006}, constituency structures continue to provide useful information regarding hierarchical organization, scope, and prosodic grouping \cite{jurafsky_speech_2026}. 

The current work adopts a constituency-oriented parsing strategy while explicitly annotating head information within grammar rules, allowing dependency relations to be derived directly from phrase structures. This design allows both representations to be preserved within a single parsing process. The main contributions of this work include an inventory of rules and lexical categories and features as well as a bottom-up parsing algorithm based on this inventory, in addition to a hybrid dependency-constituency format that the parser produces, showing full derivation for each parse, and different parse hypotheses.

\section{Related Work }

Modern syntactic parsing research is dominated by supervised statistical and neural approaches, particularly transition-based and graph-based dependency parsers, as well as neural constituency parsers. These systems typically achieve high parsing accuracy on benchmark corpora through large-scale supervised learning, typically relying on learned representations rather than manually specified grammatical constraints. Despite their empirical success, there remains continued interest in grammar-constrained and linguistically interpretable parsing systems, particularly in areas involving formal syntax, semantic compositionality, and explainable language processing.

Transition-based dependency parsers incrementally construct dependency trees through sequences of parser actions \cite{nivre_algorithms_2008}, while graph-based parsers formulate parsing as a global optimization problem over candidate dependency arcs \cite{mcdonald_non-projective_2005}. In parallel, constituency parsing has continued through neural chart parsers and span-based architectures, many of which retain bottom-up parsing principles inherited from earlier CFG and CKY-style parsing algorithms (\citealt{kasami_efficient_1966} and \citealt{younger_recognition_1967}).

Grammar-based parsing approaches based on feature-rich phrase structure formalisms remain an important line of research in computational linguistics. In particular, Generalized Phrase Structure Grammar (GPSG) and Head-driven Phrase Structure Grammar (HPSG) demonstrated that many syntactic phenomena could be represented through explicit grammatical constraints and feature propagation mechanisms without relying exclusively on transformational operations, such as movement in generative grammar.

Early computational work on GPSG and HPSG focused primarily on parsing efficiency and constraint satisfaction. For example,  \citeauthor{fisher_practical_1989} (1989) introduced parsing methods for GPSG, while \citeauthor{torisawa_hpsg_2000} (2000) proposed CFG filtering techniques to reduce the computational complexity of HPSG parsing. Subsequent work explored broad-coverage grammar engineering, including corpus-oriented Japanese HPSG parsing \cite{yoshida_corpus-oriented_2005} and hybrid constituent-dependency formulations of HPSG \cite{zhou_head-driven_2019}.

Recent work has increasingly combined symbolic grammatical formalisms with neural architectures. \citeauthor{nguyen_attempt_2024}  (2024) developed a neural parser based on simplified HPSG representations for Vietnamese, while \citeauthor{zamaraeva_revisiting_2024}  (2024) incorporated HPSG-inspired supertagging into neural constituency parsing pipelines. Similarly, \citeauthor{li_head-driven_2021} (2021) explored training HPSG parsers directly from constituency and dependency annotations. These approaches illustrate a broader trend toward neuro-symbolic parsing systems that integrate explicit grammatical structure with statistical learning methods.

A related line of work explored parsing within Minimalist Grammar and transformational frameworks. These studies investigated both formal parsing properties and probabilistic inference mechanisms for movement-based grammars, including bottom-up parsing methods \cite{harkema_recognizer_2000}, probabilistic Minimalist parsing (\citealt{mainguy_probabilistic_2010} , \citealt{portelance_grammar_2017}), and wide-coverage applications of Minimalist Grammar parsing \cite{torr_wide-coverage_2019}. While these approaches differ theoretically from GPSG and HPSG, they similarly attempt to connect formal syntactic theory with computational parsing implementations.

In parallel, deep grammar engineering projects such as the DELPH-IN ecosystem and the English Resource Grammar \footnote{https://delph-in.github.io/docs/erg/ErgTop/} demonstrated the feasibility of broad-coverage HPSG-based parsing systems using platforms such as PET \cite{callmeier_petplatform_2000}  and ACE \cite{crysmann_towards_2012}. At the same time, neural constituency parsers, including span-based and chart-based approaches (\citealt{stern_minimal_2017} and \citealt{kitaev_constituency_2018}), showed that bottom-up phrase structure parsing remains computationally effective within modern neural NLP systems, although without relying on explicit symbolic grammar rules.

Building on these approaches, the current work explores building a bottom-up parsing approach, inspired by GPSG and HPSG grammars.

\section{\textbf{Methodology} }

\subsection{\textbf{Parsing Algorithm Overview}}

The goal of this research is to develop a parsing approach capable of producing syntactically interpretable sentence structures while addressing practical challenges such as ambiguity, noisy or incomplete input, syntactic gaps, and the joint representation of constituency and dependency relations. The proposed parser follows a bottom-up incremental approach inspired by GPSG and HPSG, where syntactic structures are constructed recursively from words and their probable categories using a compact inventory of feature-based grammatical rules.

The parser proceeds left-to-right, beginning with multiple POS-tag hypotheses and their probabilistic weights. Each hypothesis is converted into an initial phrase object and recursively projected upward through unary and binary syntactic rules. Binary projections search for compatible preceding phrases while enforcing category and feature constraints. Each rule explicitly identifies the head child of the phrase, allowing dependency relations to be derived directly from constituency projections and enabling a unified syntactic representation.

Unlike transition-based and graph-based dependency parsers such as spaCy and Stanza, the current system builds syntactic structure through explicit grammatical constraints rather than parser actions or global arc optimization. The parser also incorporates weighted ambiguity resolution, limited tolerance for disfluencies and extra-textual elements, and fallback mechanisms for partially parseable input.

\begin{figure*}[htbp]
    \centering
    
    \includegraphics[width=1\linewidth]{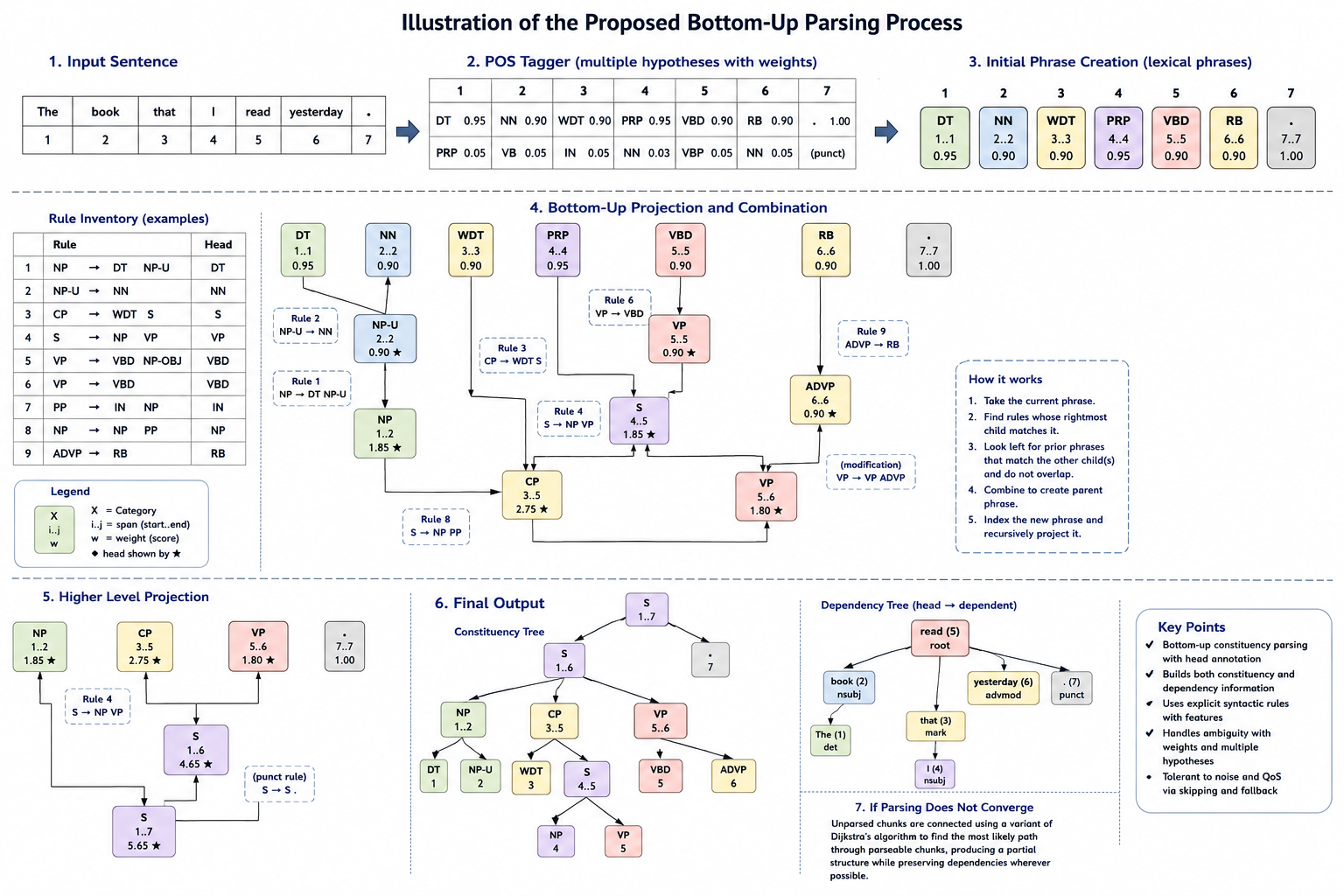}
    \caption{Parsing Algorithm Overview - Generated by ChatGPT for illustration purposes only}
    \label{fig:parsing algorithm illustration}
\end{figure*}

\subsection{\textbf{Syntactic Rules}}

One of the primary contributions of this work is the development of an extensible inventory of syntactic rules that encode grammatical structures in a transparent and maintainable format\footnote{Rule spreadsheets can be found at: https://github.com/chplu-dev/dph-parser/tree/main/rules}. These rules are inspired by principles from GPSG and HPSG while remaining computationally practical.

The rules resemble Context-Free Grammar (CFG) productions but incorporate additional feature structures and head annotations. Each rule contains one parent category and either one or two child categories. In binary rules, exactly one child is designated as the head child, enabling direct recovery of dependency relations.

A rule may be represented schematically as:

PARENT[features] → CHILD-1\textasciicircum[features] CHILD-2[features]

Categories may contain positive or negative features that constrain grammatical combinations. These features allow the grammar to capture distinctions that would otherwise be difficult to represent using ordinary CFG rules alone.

For example, noun phrases in English can recursively combine in constructions such as “United Nations Development Program.” A naive CFG rule such as:

NP → NP NP

would incorrectly license ungrammatical structures (e.g. * [NP the book the man] is good). To address this issue, the grammar introduces refined categories such as NP-U for noun phrases lacking determiners and NP-DT for noun phrases containing determiners. This permits more constrained constructions such as:

NP → NP-U NP-U and NP-DT → DT NP-U

Similarly, verb phrase structures are refined to avoid unrestricted recursion. Rather than allowing unlimited object attachment through a rule such as:

VP → VP NP

the grammar distinguishes between verb phrases containing one object and those containing two objects:

VP-O → V NP and VP-2O → VP-O NP

This constrains the grammar to structures more consistent with English syntax.

The grammar also encodes auxiliary verb sequences and feature agreement. For example, forms of the verb “to be” are projected into a BE category using feature constraints:

BE → AUX[+be]  and BE → HAVE AUX[+been]

These rules allow valid auxiliary sequences such as “would have been” while disallowing invalid combinations (e.g. * have would been).

The inventory additionally incorporates limited GPSG-style slash features for representing syntactic gaps and extraction phenomena. Head information and feature percolation mechanisms are inspired by HPSG, allowing parent categories to inherit features from their head children.

The rule inventory consists of three major components: 1) Phrase structure rules defining non-terminal projections; 2) Terminal rules mapping POS tags into syntactic categories and features; and 3) Lexical tags defining categories and features for closed-class words

The grammar is intentionally extensible and maintainable through spreadsheet-style representations and version-controlled rule inventories. The purpose of the current work is not to exhaustively define all grammatical structures of English, but rather to demonstrate a flexible methodology for integrating linguistic theory into practical parsing systems.

\subsection{\textbf{POS Tagging}}

The parser depends on probabilistic POS tagging as the initial stage of the parsing pipeline. Although many POS taggers are available, the current system requires specific properties: 1) The tag set must align with the categories used in the syntactic rules; 2) The tagger must provide confidence scores or probabilities; and 3) The tagger should produce multiple tag hypotheses for each token.

Producing multiple hypotheses is particularly important because the most probable POS tag is not always the tag that yields the best global parse.

To satisfy these requirements, a simplified POS tagger based on Recurrent Neural Networks (RNNs) was developed. The input consists of sequences of tokens, while the output is a probability distribution over both Universal POS (UPOS) and language-specific XPOS tags from the Universal Dependencies framework. XPOS tags are preferred over generic UPOS tags because they provide finer-grained distinctions necessary for syntactic rules. For example, different rules apply to gerunds, infinitives, participles, and finite verbs. 

The features used by the POS tagger include leading and trailing characters within each token, capitalization patterns, and orthographic properties such as hyphenation. The model was trained using training data from Universal Dependencies treebanks. Parameters: LSTM network - hidden\_size=64, n\_layers=3, learning rate=0.0001, Adam Optimizer, Cross Entropy Loss, applying sigmoid at the output. This tagger learns from training data, therefore it is the only supervised part of the current study. 

The output of the POS tagger forms the initial set of weighted terminal phrase objects used by the parser.

\subsection{Lexical Tags}

In addition to the output of POS tagger, and as part of the rule inventory, we introduce a custom list for lexical items and their categories and features. This list of lexical tags is mainly for function words, or words with a specific grammatical role. Such tags are not typically captured by standard UPOS or XPOS tags. For example, the word “is” takes “AUX” and “VBZ” tags, although it has specific grammatical properties separating it from other AUX and Verb tags. Therefore, we assign a custom category “BE” to verb to-be words (is, was, were, am … etc). A similar situation applies for the verb to-have and to-do. Additional features are also introduced to identify how for example personal pronouns can be used as determiners before nouns (e.g. my book - the book). This custom lexical information augments the output of the POS tagger, while matching categories and features used in the syntactic rules. 

\subsection{\textbf{Phrase Representation}}

The central data structure in the parser is the phrase object. Each phrase represents either a terminal lexical category or a non-terminal syntactic constituent.

Each phrase contains the following information:

[Phrase start position, Phrase end position, Syntactic category, Feature set, Unique phrase identifier, Child phrase identifiers, Head child information, Weight or confidence score, Dependency relation information]

These phrase objects form the basis for all subsequent parsing operations.

\subsection{\textbf{Parsing and Phrase Projection}}

The parsing process begins by converting each POS-tag hypothesis, as well as those produced from lexical tags, into an initial phrase object. Each phrase is then recursively projected using the syntactic rule inventory.

Projection refers to the process of identifying rules whose child categories match the current phrase. Unary rules involve projection from one category into another (e.g. NP → NN), while binary rules combine two adjacent phrases into a larger constituent. For binary rules, the parser first identifies rules whose final child matches the current phrase. It then searches previously created phrases to locate a compatible preceding phrase that satisfies the first child of the rule. For example, given the rule:

PP → IN NP

if the current phrase is NP, the parser scans preceding phrases to identify a compatible IN phrase. If found, the two phrases are combined into a PP phrase.

When phrases are combined, the parser creates a new parent phrase with updated information from the two child phrases, including the total weight of both, combined span, and which child is the head. The resulting phrase is indexed and recursively projected further.

\subsection{\textbf{Parser Indexing and Optimization}}

Efficient indexing is essential because parsing generates many overlapping phrase hypotheses. The parser therefore maintains several indexing structures for rapid lookup of phrases and rules. The primary indexing structure organizes phrases according to their spans and categories:

Dictionary[end][start][category] = [phrase1, phrase2, ...]

Additional indexes are maintained for quickly retrieving rules associated with particular categories and features. These indexing strategies substantially reduce the search space and improve parsing efficiency.

\subsection{\textbf{Noise Handling and Fallback Parsing}}

Natural language input frequently contains disfluencies, interruptions, punctuation artifacts, or incomplete constructions. To improve robustness, the parser allows limited skipping between adjacent phrases during binary matching. This permits phrases to combine even when separated by small amounts of noise. Nevertheless, some sentences may remain partially unparsable because of grammatical irregularities or missing rules. In such cases, the parser applies a fallback mechanism inspired by Dijkstra’s shortest-path algorithm. Rather than forcing a fully convergent parse, the system identifies the most likely sequence of parseable chunks and connects them into a partial structure.  

\subsection{\textbf{Output Representation and Evaluation}}

The final parser output includes both constituency and dependency representations. The highest-scoring phrase spanning the full sentence, or the largest parseable span, is selected as the primary parse. Recursive traversal of phrase children produces:

\begin{itemize}
    \item A constituency tree representation (converted to PTB bracketed format)
    \item Dependency relations between words (converted to CoNLL tabular format)
\end{itemize}
The parser is evaluated both qualitatively and quantitatively. Qualitative evaluation involves manual inspection of parse outputs and their consistency with predefined syntactic rules. Quantitative evaluation involves measuring parsing performance on multiple corpora and comparing the results against widely used systems such as spaCy and Stanza.

\section{Experimental Setup }

\subsection{Metrics}

In order to evaluate the performance of the current parsing approach, we use Unlabelled Attachment Score (UAS). This metric reveals the quality of parsing mainly in terms of showing which words depend on which.  The measurement counts how many tokens the parser identified their heads correctly, divided by the total number of tokens in the dataset. 

\subsection{Data}

This parser is tested on the English dev/test datasets from Universal Dependencies (UD) \footnote{\href{https://universaldependencies.org/}{https://universaldependencies.org} }  \cite{nivre_universal_2020}. It includes multiple corpora with different distributions for train/dev/test sets. It includes 8 files in the train set, 7 files in the dev set, and 13 files for the test sets. Each file includes sentences annotated according to the CoNLL format for dependency parsing. This parser is applied on the tokenized sentences in each CoNLL item from the treebank. 

For benchmarking, we use spaCy dependency parser\footnote{\href{https://spacy.io/api/dependencyparser}{https://spacy.io/api/dependencyparser}  }, transition based parser \cite{honnibal_joint_2014} . We also use Stanza \footnote{\href{https://stanfordnlp.github.io/stanza/}{https://stanfordnlp.github.io/stanza/} }, a graph-based parser \cite{qi_stanza_2020} . The results are based on the standard models for both spaCy and Stanza without any training or fine-tuning, by applying it to the tokenized sentences from the treebank. We use the UD evaluation tools \footnote{\href{https://universaldependencies.org/tools.html}{https://universaldependencies.org/tools.html}  } to measure the UAS of produced parses in CoNLL format against the Gold-Standard parses of the respective corpora. 

\subsection{Code}

The code used for these experiments, including notebooks, POS models, python files, Excel sheets for rules and other lists, is included in a GitHub repository for easy access and reproducibility \footnote{\href{https://github.com/chplu-dev/dph-parser/}{https://github.com/chplu-dev/dph-parser/} } . The readme file includes relevant instructions for using the code.

\section{Results and Discussion }
\label{sec:bibtex}
\begin{figure*}
    \centering
    \includegraphics[width=1\linewidth]{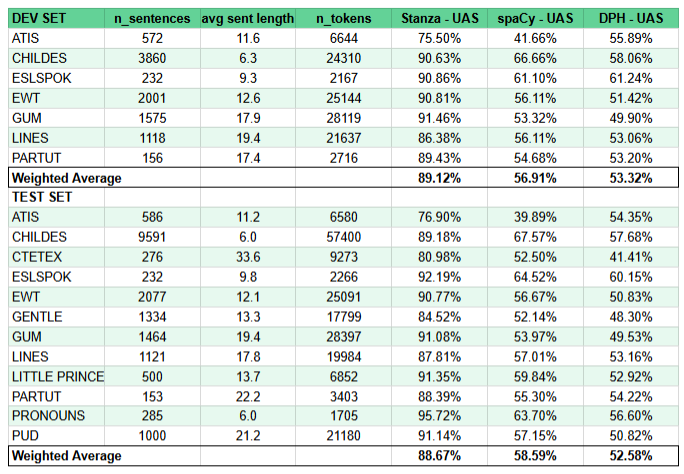}
    \caption{UAS results from DPH parser against spaCy and Stanza }
    \label{fig:main-results-table}
\end{figure*}

For quantitative evaluation, we compare the UAS metrics from DPH parser to those from the spaCy and Stanza parsers. Results are obtained on Development/Test sections of UD datasets (devset/testset). The performance figures reflect how many phrases and dependencies are identified correctly by the parser. The average UAS for each parser across devset/testset is weighted according to the number of tokens within each dataset.  

As we can see from figure 3, DPH parser achieved UAS of 53.32\% \& 52.58\% (UD Devset/Testset respectively). For the same data, Stanza achieved 89.12\% \& 88.67\% while spaCy achieved 56.91\% and 58.59\%. The performance of spaCy is lower than expected, given how commonly used it is, also given that UAS values for UD datasets for the shared task in 2018 reached up to 80.51\% \footnote{https://universaldependencies.org/conll18/results-uas.html}.  However, it should be noted that we used the off-the-shelf smallest model, without training on the UD training data or any other fine tuning. 

It is also noteworthy that spaCy performed much worse on ATIS dataset \footnote{https://github.com/UniversalDependencies/UD\_English-Atis} (UAS devset: 41.66\% - testet: 39.89\%). This dataset warrants further analysis, since also Stanza had the lowest UAS values at this dataset (devset 75.50\%, testset: 76.90\%). On the other hand, DPH parser achieved average performance for this dataset (devset 55.89\%, testset 54.35\%).     

Although the performance of DPH parser is much lower than Stanza, it is only a few points below spaCy performance. In addition, we can see that the DPH performance is mostly robust across multiple datasets, even surpassing spaCy in multiple datasets across the devset and testset, as we can see from the ATIS example. 

At this stage, the parser is only programmed with a small subset of rules (148 rules) of the English language, just as a proof of concept. While these rules do not provide full coverage of English grammar, it shows proof of concept that programming even a small subset can be used for parsing various datasets. Further investigation is needed to measure the effect of wider coverage in improving parsing outcomes.

\begin{figure*}
    \centering
    \includegraphics[width=1\linewidth]{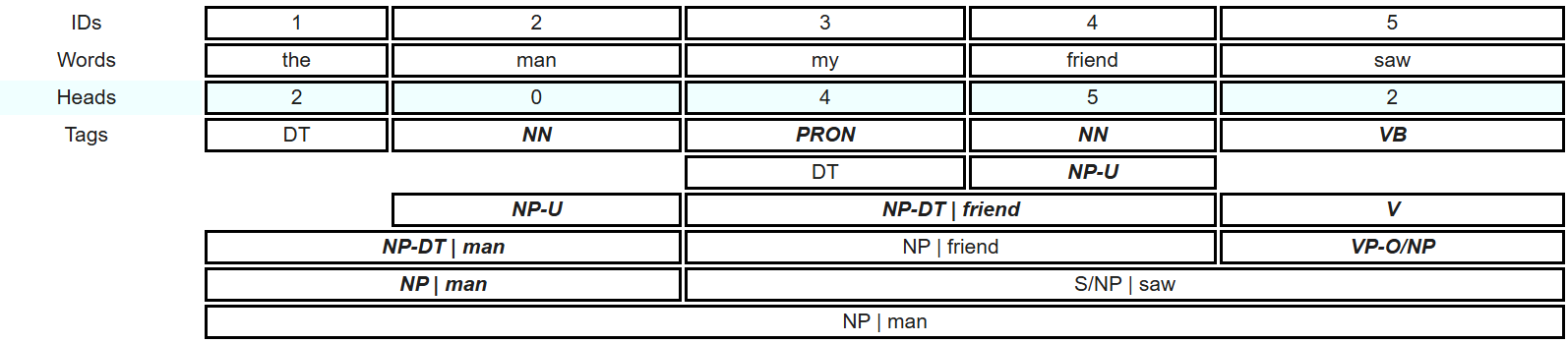}
    \caption{Slash representation derivation for VP with missing NP}
    \label{fig:VP-NP-Slash}
\end{figure*}

Qualitatively, we notice that parser output reflects both constituency and dependency structures. The representation proposed includes the tokens of the input sentence, as well as their IDs, heads and tags, which captures some of the main information from the dependency structure. It also includes a hierarchical listing of the phrases, their levels and their spans, as we can see in figure 4.

An important point to examine is the applicability of this parsing approach to custom syntactic rules pertaining to certain linguistic structures and phenomena. For this case, we address, as an example, the slash features within GPSG. We can see in figure 4 that the parser was able to correctly handle and represent the structure of the phrase: 

"the man my friend saw X" 

The output shows the derivation of all the rules applied. Such rules include the slash rule VP-O/NP corresponding to a VP missing a direct object (the gap construction), and its projection into S/NP. Eventually, we see the application of the rule to fill the gap:

NP → NP\textasciicircum  \  S/NP

We can apply similar rules for Prepositional Phrases missing a noun phrase and WH-questions.

\section{Conclusions}
In this study, we revisited grammar-based parsing, and examined whether it can provide new insights into the modern language processing landscape. For this purpose, we explored the possibility of using an inventory of explicit syntactic rules, as well as listings of lexical categories and features for closed-set words, as the basis for a bottom-up parsing approach. The pipeline built through the syntactic rule inventory and the bottom-up parsing algorithm was shown to be capable of parsing real-world data, with initially promising performance. It is important to note that the goal of this approach, at this stage, is not competitive parsing. However, we report its parsing outcomes, measured by UAS metrics as the baseline for future implementations.   

Another important goal of this study is to leverage some of the knowledge acquired over decades in the field of syntax and other areas of theoretical linguistics. It is important to be able to frame this knowledge into computational frameworks, usable for NLP applications. We started addressing this goal as part of our qualitative evaluation, by feeding sentences exhibiting specific syntactic phenomena to the parser and evaluating its output accordingly. As an example, we attempted to represent the syntactic gap construction among the syntactic rules processed by the DPH parser, using GPSG slash notation. The qualitative results indicated that the parser can indeed process such constructions with the expected derivation. However, more detailed quantitative investigation is needed for how accurately the parser captures these constructions in real-world data.  

This parsing approach shows the full grammatical derivation used to achieve the final structure. In fact, it produces multiple parse hypotheses, both for the sentence as a whole, and for each chunk of words. This will allow automatic detection of syntactic ambiguity and where there can be multiple meanings. It will also allow re-ranking these hypotheses to identify the most likely parse. In addition, this approach produces a richer combination of dependency and constituency structures, leveraging the strengths of both paradigms. It is compatible with modern parsing tools and evaluation schemes and metrics. 

With this new approach, it is possible to expand to any new language. What is needed is the following: 1) An inventory of possible POS tags, lexical categories and features for this language; 2) An inventory of GPSG/HPSG rules for this language; 3) A framework for predicting POS tags for this language; and 4) For languages with rich morphology, also a framework for predicting the likely sub-tokenization of prefixes, suffixes and subwords

This work thus contributes to ongoing efforts to integrate linguistic formalisms with empirically testable computational modeling. Further, it allows more explainable parsing behavior, including identifying syntactic ambiguities and allowing closer inspection of both underlying constituency and dependency structures. Finally, it allows building language-specific rule inventory, to leverage theoretical syntactic knowledge possibly into low-resource and understudied languages. 

\section*{Limitations}

The results discussed in this paper are only related to English. The rules and features used in the system are hand crafted only as a sample and do not represent a comprehensive listing of English syntactic rules. The Parts of Speech tagger included was trained on training data from Universal Dependencies. All fine tuning of rules and algorithms was according to the UD training data. Results only pertain to corpora included in UD dev/test datasets. 

\section*{Acknowledgments}


\bibliography{references}

@inproceedings{crysmann_towards_2012,
	title = {Towards efficient {HPSG} generation for {German}, a non-configurational language},
	booktitle = {Proceedings of {COLING} 2012},
	author = {Crysmann, Berthold and Packard, Woodley},
	year = {2012},
	pages = {695--710},
}

@article{callmeier_petplatform_2000,
	title = {{PET}–a platform for experimentation with efficient {HPSG} processing techniques},
	volume = {6},
	number = {1},
	journal = {Natural Language Engineering},
	publisher = {Cambridge University Press},
	author = {Callmeier, Ulrich},
	year = {2000},
	pages = {99--107},
}

@article{younger_recognition_1967,
	title = {Recognition and parsing of context-free languages in time n3},
	volume = {10},
	number = {2},
	journal = {Information and control},
	publisher = {Elsevier},
	author = {Younger, Daniel H},
	year = {1967},
	pages = {189--208},
}

@article{kasami_efficient_1966,
	title = {An efficient recognition and syntax-analysis algorithm for context-free languages},
	journal = {Coordinated Science Laboratory Report no. R-257},
	publisher = {Coordinated Science Laboratory, University of Illinois at Urbana-Champaign},
	author = {Kasami, Tadao},
	year = {1966},
}

@inproceedings{kitaev_constituency_2018,
	title = {Constituency parsing with a self-attentive encoder},
	booktitle = {Proceedings of the 56th {Annual} {Meeting} of the {Association} for {Computational} {Linguistics} ({Volume} 1: {Long} {Papers})},
	author = {Kitaev, Nikita and Klein, Dan},
	year = {2018},
	pages = {2676--2686},
}

@inproceedings{stern_minimal_2017,
	title = {A minimal span-based neural constituency parser},
	booktitle = {Proceedings of the 55th {Annual} {Meeting} of the {Association} for {Computational} {Linguistics} ({Volume} 1: {Long} {Papers})},
	author = {Stern, Mitchell and Andreas, Jacob and Klein, Dan},
	year = {2017},
	pages = {818--827},
}

@inproceedings{mcdonald_non-projective_2005,
	title = {Non-projective dependency parsing using spanning tree algorithms},
	booktitle = {Proceedings of human language technology conference and conference on empirical methods in natural language processing},
	author = {McDonald, Ryan and Pereira, Fernando and Ribarov, Kiril and Hajic, Jan},
	year = {2005},
	pages = {523--530},
}

@article{nivre_algorithms_2008,
	title = {Algorithms for deterministic incremental dependency parsing},
	volume = {34},
	number = {4},
	journal = {Computational Linguistics},
	author = {Nivre, Joakim},
	year = {2008},
	pages = {513--553},
}

@book{jurafsky_speech_2026,
	edition = {3rd},
	title = {Speech and {Language} {Processing}: {An} {Introduction} to {Natural} {Language} {Processing}, {Computational} {Linguistics}, and {Speech} {Recognition}, with {Language} {Models}},
	url = {https://web.stanford.edu/~jurafsky/slp3/},
	author = {Jurafsky, Daniel and Martin, James H.},
	year = {2026},
}

@inproceedings{qi_stanza_2020,
	title = {Stanza: {A} {Python} natural language processing toolkit for many human languages},
	booktitle = {Proceedings of the 58th annual meeting of the association for computational linguistics: system demonstrations},
	author = {Qi, Peng and Zhang, Yuhao and Zhang, Yuhui and Bolton, Jason and Manning, Christopher D.},
	year = {2020},
	pages = {101--108},
}

@article{torisawa_hpsg_2000,
	title = {An {HPSG} parser with {CFG} filtering},
	volume = {6},
	number = {1},
	journal = {Natural Language Engineering},
	publisher = {Cambridge University Press},
	author = {Torisawa, Kentaro and Nishida, Kenji and Miyao, Yusuke and Tsujii, Jun-Ichi},
	year = {2000},
	note = {ISBN: 1469-8110},
	pages = {63--80},
}

@book{chomsky_syntactic_1957,
	title = {Syntactic structures},
	isbn = {3-11-017279-8},
	publisher = {Walter de Gruyter},
	author = {Chomsky, Noam},
	year = {1957},
}

@inproceedings{zamaraeva_revisiting_2024,
	title = {Revisiting {Supertagging} for faster {HPSG} parsing},
	booktitle = {In {Proceedings} of the 2024 {Conference} on {Empirical} {Methods} in {Natural} {Language} {Processing}, pp. 11359-11374.},
	author = {Zamaraeva, Olga and Gómez-Rodríguez, Carlos},
	year = {2024},
}

@inproceedings{yoshida_corpus-oriented_2005,
	title = {Corpus-oriented development of japanese {HPSG} parsers},
	booktitle = {In {Proceedings} of the {ACL} {Student} {Research} {Workshop}, pp. 139-144.},
	author = {Yoshida, Kazuhiro, Yoshida},
	year = {2005},
}

@book{nivre_inductive_2006,
	title = {Inductive dependency parsing},
	publisher = {Springer},
	author = {Nivre, Joakim},
	year = {2006},
}

@article{xia_converting_2001,
	title = {Converting dependency structures to phrase structures},
	author = {Xia, Fei and Palmer, Martha},
	year = {2001},
}

@inproceedings{torr_wide-coverage_2019,
	title = {Wide-coverage neural {A}* parsing for minimalist grammars},
	booktitle = {Proceedings of the 57th annual meeting of the association for computational linguistics},
	author = {Torr, John and Stanojević, Miloš and Steedman, Mark and Cohen, Shay B.},
	year = {2019},
	pages = {2486--2505},
}

@article{portelance_grammar_2017,
	title = {Grammar {Induction} for {Minimalist} {Grammars} using {Variational} {Bayesian} {Inference}: {A} {Technical} {Report}},
	journal = {arXiv preprint arXiv:1710.11350},
	author = {Portelance, Eva and Bruno, Amelia and Harasim, Daniel and Bergen, Leon and O'Donnell, Timothy J.},
	year = {2017},
}

@inproceedings{nivre_universal_2020,
	title = {Universal {Dependencies} v2: {An} evergrowing multilingual treebank collection},
	booktitle = {Proceedings of the twelfth language resources and evaluation conference},
	author = {Nivre, Joakim and De Marneffe, Marie-Catherine and Ginter, Filip and Hajic, Jan and Manning, Christopher D. and Pyysalo, Sampo and Schuster, Sebastian and Tyers, Francis and Zeman, Daniel},
	year = {2020},
	pages = {4034--4043},
}

@article{zhou_head-driven_2019,
	title = {Head-driven phrase structure grammar parsing on {Penn} treebank},
	journal = {arXiv preprint arXiv:1907.02684},
	author = {Zhou, Junru and Zhao, Hai},
	year = {2019},
}

@inproceedings{nguyen_attempt_2024,
	title = {An {Attempt} to {Develop} a {Neural} {Parser} based on {Simplified} {Head}-{Driven} {Phrase} {Structure} {Grammar} on {Vietnamese}},
	booktitle = {International {Symposium} on {Information} and {Communication} {Technology}},
	publisher = {Springer},
	author = {Nguyen, Duc-Vu and Phan, Thang Chau and Nguyen, Quoc-Nam and Nguyen, Kiet Van and Nguyen, Ngan Luu-Thuy},
	year = {2024},
	pages = {313--328},
}

@article{mainguy_probabilistic_2010,
	title = {A probabilistic top-down parser for minimalist grammars},
	journal = {arXiv preprint arXiv:1010.1826},
	author = {Mainguy, Thomas},
	year = {2010},
}

@article{li_head-driven_2021,
	title = {Head-driven {Phrase} {Structure} {Parsing} in {O} (\$ n{\textasciicircum} 3\$) {Time} {Complexity}},
	journal = {arXiv preprint arXiv:2105.09835},
	author = {Li, Zuchao and Zhou, Junru and Zhao, Hai and Parnow, Kevin},
	year = {2021},
}

@article{honnibal_joint_2014,
	title = {Joint incremental disfluency detection and dependency parsing},
	volume = {2},
	journal = {Transactions of the Association for Computational Linguistics},
	publisher = {MIT Press One Rogers Street, Cambridge, MA 02142-1209, USA journals-info …},
	author = {Honnibal, Matthew and Johnson, Mark},
	year = {2014},
	note = {ISBN: 2307-387X},
	pages = {131--142},
}

@inproceedings{harkema_recognizer_2000,
	title = {A recognizer for minimalist grammars},
	booktitle = {Proceedings of the {Sixth} {International} {Workshop} on {Parsing} {Technologies}},
	author = {Harkema, Henk},
	year = {2000},
	pages = {111--122},
}

@article{fisher_practical_1989,
	title = {Practical parsing of generalized phrase structure grammars},
	volume = {15},
	number = {3},
	journal = {Computational Linguistics},
	author = {Fisher, Anthony J.},
	year = {1989},
	pages = {139--148},
}

@book{pollard_head-driven_1994,
	title = {Head-driven phrase structure grammar},
	isbn = {0-226-67447-9},
	publisher = {University of Chicago Press},
	author = {Pollard, Carl and Sag, Ivan A.},
	year = {1994},
}

@book{gazdar_generalized_1985,
	title = {Generalized phrase structure grammar},
	isbn = {0-674-34455-3},
	publisher = {Harvard University Press},
	author = {Gazdar, Gerald},
	year = {1985},
}

\appendix

\section{Appendix}
\label{sec:appendix}

\end{document}